\documentclass[conference, a4paper]{IEEEtran}
\IEEEoverridecommandlockouts
\usepackage[top=0.70in,bottom=1in,left=0.62in,right=0.62in]{geometry}
\usepackage{cite}
\usepackage{amsmath,amssymb,amsfonts}
\usepackage{graphicx}
\usepackage{subfigure}
\usepackage{textcomp}
\usepackage{xcolor}
\usepackage{eso-pic}
\usepackage{titlesec}
\usepackage{url}
\usepackage{multicol}
\usepackage{multirow}
\usepackage{amsmath}
\usepackage{algpseudocode}
\usepackage{lipsum}
\usepackage{fancybox}
\usepackage{listings}
\usepackage[utf8]{inputenc}
\usepackage{comment}
\usepackage{hyperref}
\usepackage{url}
\usepackage{adjustbox}
\usepackage{array}
\usepackage{fancyhdr}
\usepackage[ruled,vlined]{algorithm2e}
\SetKwComment{Comment}{$\triangleright$\ }{}

\titleformat{\subsection}{\normalsize\bfseries}{\thesubsection}{1em}{}

\def\BibTeX{{\rm B\kern-.05em{\sc i\kern-.025em b}\kern-.08em
    T\kern-.1667em\lower.7ex\hbox{E}\kern-.125emX}}

\def\thesection{\arabic{section}}  % Set chapters/sections to numeric (1, 2, 3...)
\def\thesubsection{\thesection.\arabic{subsection}}
\def\thesubsubsection{\thesubsection.\arabic{subsubsection}}

\def\thesectiondis{\thesection}                     % Numeric section heading
\def\thesubsectiondis{\thesectiondis.\arabic{subsection}}
\def\thesubsubsectiondis{\thesubsectiondis.\arabic{subsubsection}}

\makeatletter

\renewcommand\section{\@startsection {section}{1}{\z@}% 
    {-3.5ex \@plus -1ex \@minus -.2ex}% Adjusts spacing before
    {2.3ex \@plus.2ex}% Adjusts spacing after
    {\normalfont\normalsize\bfseries\raggedright}} % Left-align section titles

\def\@makechapterhead#1{%
    \vspace*{50\p@}% Vertical space before title
    {\parindent \z@ \raggedright \normalfont
     \huge\bfseries \thechapter\hspace{1em}#1\par\nobreak
     \vskip 40\p@}}

\def\ps@IEEEtitlepagestyle{%
    \def\@oddfoot{\mycopyrightnotice}%
    \def\@evenfoot{}%
}

\def\mycopyrightnotice{
  {\footnotesize ~\copyright2025 ~IEEE. This paper has been accepted at the IEEE QPAIN 2025. The final version will be available in the IEEE Xplore Digital Library.\hfill} % <--- Change here
  \gdef\mycopyrightnotice{}
}

\makeatother
\def\confheader#1{%
    \def\ps@IEEEtitlepagestyle{%
        \old@ps@IEEEtitlepagestyle%
        \def\@oddhead{\strut\hfill#1\hfill\strut}%
        \def\@evenhead{\strut\hfill#1\hfill\strut}%
    }%
    \ps@headings%
}
\makeatother
\begin{document}

\newcommand\AtPageUpperMyright[1]{\AtPageUpperLeft{
 \put(\LenToUnit{0.06\paperwidth},\LenToUnit{-1cm}){ % moved far left
     \parbox{0.78\textwidth}{\raggedright\fontsize{9}{11}\selectfont #1}} % left-aligned text
 }}

\title{Jellyfish Species Identification: A CNN Based Artificial Neural Network Approach \vspace{-1mm}\\
%\conf{International Conference on Quantum Photonics, Artificial Intelligence, and Networking (QPAIN) \\ 31 July - 2 August 2025, Rangpur, Bangladesh}
\thanks{\vspace{10pt}\noindent\parbox[b]{\dimexpr\columnwidth-1em}{ % Restrict to left column
\textbf{\rule{5.1cm}{1pt}}\\ % Custom line width (5cm) and thickness (1.5pt)
*Corresponding Author: Pabon Shaha \vspace{2mm} \\  This research is supported by "Bangladesh University and NextGen AI Lab"; we are grateful for the resource, guidance, and encouragement provided by the institution. }}
}

\author{\IEEEauthorblockN{Md. Sabbir Hossen\IEEEauthorrefmark{9}, Md. Saiduzzaman\IEEEauthorrefmark{9}, Pabon Shaha\IEEEauthorrefmark{6}\IEEEauthorrefmark{1}, and Mostofa Kamal Nasir\IEEEauthorrefmark{6}}
\IEEEauthorblockA{\IEEEauthorrefmark{9}Dept. of Computer Science \& Engineering, Bangladesh University, Dhaka, Bangladesh}
\IEEEauthorblockA{\IEEEauthorrefmark{6}Dept. of Computer Science \& Engineering, Mawlana Bhashani Science \& Technology University, Tangail, Bangladesh}
\IEEEauthorblockA{Email: sabbir.hossen@bu.edu.bd, saiduzzamancse56bu@gmail.com, pabonshahacse15@gmail.com, kamal@mbstu.ac.bd }
}

\maketitle

\begin{abstract}
Jellyfish, a diverse group of gelatinous marine organisms, play a crucial role in maintaining marine ecosystems but pose significant challenges for biodiversity and conservation due to their rapid proliferation and ecological impact. Accurate identification of jellyfish species is essential for ecological monitoring and management. In this study, we proposed a deep learning framework for jellyfish species detection and classification using an underwater image dataset. The framework integrates advanced feature extraction techniques, including MobileNetV3, ResNet50, EfficientNetV2-B0, and VGG16, combined with seven traditional machine learning classifiers and three Feedforward Neural Network classifiers for precise species identification. Additionally, we activated the softmax function to directly classify jellyfish species using the convolutional neural network models. The combination of the Artificial Neural Network with MobileNetV3 is our best-performing model, achieved an exceptional accuracy of 98\%, significantly outperforming other feature extractor-classifier combinations. This study demonstrates the efficacy of deep learning and hybrid frameworks in addressing biodiversity challenges and advancing species detection in marine environments.
 
 \vspace{2mm}
\end{abstract}

\begin{IEEEkeywords}
Machine Learning, Deep Learning, Multilayer Perceptron, Underwater Object Detection, Jellyfish Detection \vspace{-2mm}
\end{IEEEkeywords}

\section{INTRODUCTION}
\vspace{-2mm}
Jellyfish, belonging to the Phylum Cnidaria, are gelatinous marine organisms found in oceans worldwide \cite{phylum}. These creatures are vital to the marine ecosystem, serving as both predators and prey, but they can also pose significant challenges to human activities and marine biodiversity \cite{ecological, eco}. While jellyfish blooms contribute to nutrient cycling and provide a food source for marine species, they can cause harm by stinging swimmers, clogging power plant cooling systems, and disrupting fishing operations \cite{harmful}. Accurate identification of jellyfish species is crucial for ecological management, monitoring biodiversity, and mitigating the adverse effects of jellyfish blooms. Traditional methods of species identification are time-consuming and prone to errors, especially when dealing with underwater images affected by environmental factors such as low lighting and turbidity \cite{iden}. Recent advancements in computer vision and machine learning offer promising solutions to automate and enhance jellyfish species detection and classification \cite{mliden}.  This study introduces a deep learning framework to classify jellyfish species from underwater images. The dataset used in this research comprises thousands of underwater jellyfish images collected from publicly available sources and dedicated marine research organizations. The proposed framework employs state-of-the-art CNN architectures to directly classify and extract features, including MobileNetV3, ResNet50, EfficientNetV2-B0, and VGG16. A hybrid approach was also implemented by integrating the extracted features and traditional machine learning classifiers alongside three Feedforward Neural Network classifiers to ensure robust and accurate performance. The models were evaluated using standard metrics such as accuracy, precision, recall, and F1 score. The remainder of this paper is structured as follows. Section II reviews existing literature on jellyfish and underwater object detection and its ecological significance. Section III describes the dataset preparation, preprocessing steps, and the proposed framework and model architecture. Section IV presents the experimental setup and Implementation. Section V shows the experimental results and their comparative analysis. Finally, Section VI concludes with a summary of findings and directions for future research in jellyfish detection and ecological monitoring.

%According to a 2022 report by the Marine Biological Association, the global frequency of jellyfish blooms has increased significantly due to climate change, overfishing, and pollution. For instance, in the Mediterranean Sea alone, jellyfish blooms have led to an estimated annual economic loss of €350 million in the fisheries and tourism industries.

\section{LITERATURE REVIEW}
\vspace{-2mm} The field of jellyfish classification and detection is significantly underdeveloped, with limited studies addressing this area. Challenges like poor visibility and noise in underwater imagery make it difficult to develop effective models. To overcome this, researchers often draw from related fields like underwater object detection, including coral reef monitoring, fish detection, and marine debris classification. These advancements provide useful techniques that can be adapted for jellyfish classification. This chapter explores the limited work on jellyfish detection alongside relevant advancements in underwater object detection to frame the context and underline the necessity of further research in this domain. \textit{M. Vishwakarma et al.} \cite{vishwakarma2024jellyfish} used the CNN architecture for deep learning-based feature extraction from images, utilizing transfer learning with the MobileNetV2 architecture, and offers a thorough method for jellyfish detection. A variety of jellyfish species are used to train the CNN model. In parallel, analyzing picture characteristics retrieved by resizing and flattening uses a conventional machine learning model, SVM. Three different kinds of jellyfish make up the dataset used to train the SVM model. Results from experiments show that both models were successful; the CNN achieved a high accuracy of 97\% on the training dataset, while the SVM performed well on a different test set. \textit{G.M Firdaus et al.} 
\cite{firdaus2024improved} tackles the problem of jellyfish image classification by utilizing the EfficientNetB3 architecture and the power of Deep Convolutional Neural Networks (DCNNs). This study used a dataset of 900 jellyfish photos from 6 classes and the EfficientNetB3 model, which was selected for its harmony between complexity and performance efficiency. The EfficientNetB3 DCNNs beat traditional CNNs in image classification tasks with an accuracy of 96.67\%, according to an evaluation of the model's performance based on classification accuracy. \textit{U. Nawarathne et al.} \cite{nawarathne2024comparative} explored a variety of deep learning models for jellyfish classification, including the newly introduced YOLO architecture, YOLOv8, and more conventional convolutional neural networks (CNNs), including DenseNet, InceptionV3, MobileNet, MobileNetV2, and NASNet Mobile. With an astounding 99.5\% accuracy rate, Yolov8 stands out as particularly effective at precisely identifying and categorizing jellyfish occurrences.  \textit{T.-N. Pham et al.} \cite{pham2024improved} developed a jellyfish detection model using CNN-based deep-learning object identification models using optical data. This paper introduces an improved YOLOv5- nano model based on the addition of GAM and the substitution of CoordCov modules for traditional Conv modules in the backbone of the conventional structure.  The experiment results demonstrate that the suggested enhanced model outperforms the others, with an accuracy of 89.1\% mAP@0.5, surpassing RetinaNet, SSD, Faster R-CNN, YOLOv6, and YOLOv8. \textit{Y. Han et al.} \cite{han2022research} utilized deep learning methodology to classify different types of jellyfish. The authors used several deep learning techniques, including Google and AlexNet. In their experiment, the accuracy of the jellyfish classification job based on the GoogLeNet backbone network is 96.21\%, which is higher than AlexNet, according to a comparison of the classification results from the two networks. Lastly, the two backbone networks were utilized to examine the detection performance of the Faster R-CNN algorithm, which is employed to identify jellyfish. With an average detection accuracy of 74.96\%, the findings demonstrate that the Faster R-CNN method, based on GoogLeNet, has a greater detection accuracy in the jellyfish identification test. \textit{S Roy et al.} \cite{10829461} developed a reliable machine learning-based underwater object classification system to identify items in underwater images.  This study examines the effectiveness of transfer learning with VGG16. Pre-trained weights that provide information on basic picture attributes are usefully provided by VGG16. Using three datasets, they trained a convolutional neural network and a VGG-16-based neural network classifier. Using transfer learning, the VGG-16-based classifier achieved an accuracy of 84.89\%. \textit{J. Roy et al.} \cite{roy2024under} used the YOLOv8 models, enhanced with the MaxRGB filter, to reliably recognize and categorize underwater objects. To improve the content separability of photos, they employ the MaxRGB filter in their experiment. The suggested approach improves classification accuracy by 99.6\% and archives 98.6\% mean average precision.

\section{MATERIALS \& METHODS} 
\begin{figure}[htbp]
	\centering 
	\fbox{\includegraphics[height=11cm, width=8.5cm]{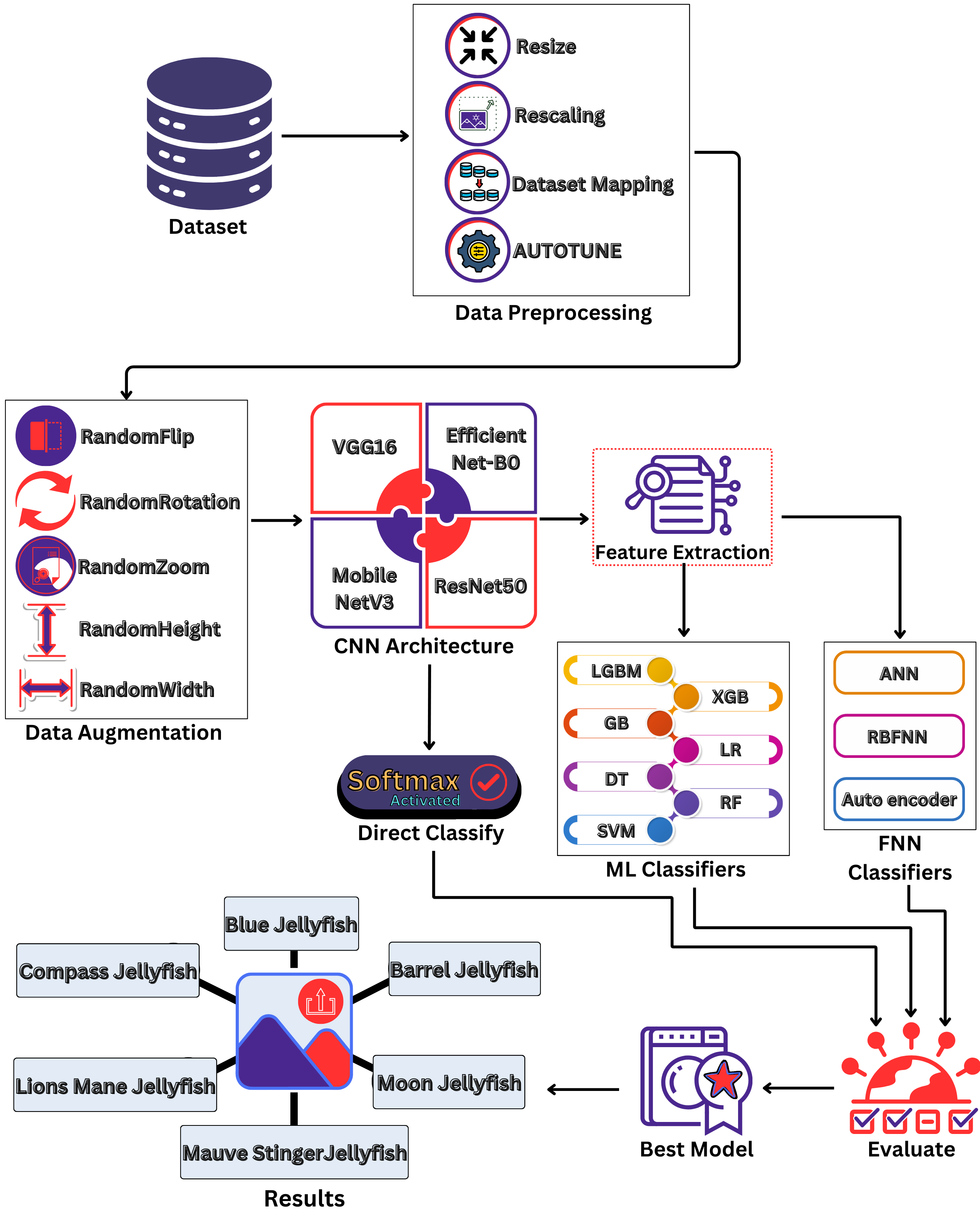}}
		\caption{Proposed Workflow Diagram for Jellyfish Species Identification} 
	\label{fig:pmd}
\end{figure}

The method used in this research is covered in this section of our paper. Dataset description, preprocessing, data augmentation, feature extraction, machine learning and multilayer perceptron model selection are some of the steps that are included in this process. Figure \ref{fig:pmd} shows the proposed workflow diagram. \vspace{-2mm}
 \vspace{-2mm}
\begin{figure*}[htbp]
    \centering
    \subfigure[Barrel]{\includegraphics[height=2cm, width=2.87cm]{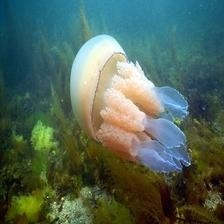}}
    \subfigure[Barrel]{\includegraphics[height=2cm, width=2.87cm]{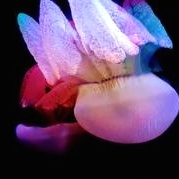}}
    \subfigure[Barrel]{\includegraphics[height=2cm, width=2.87cm]{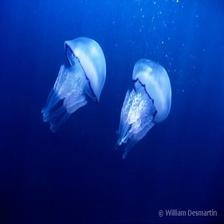}}
    \subfigure[Blue]{\includegraphics[height=2cm, width=2.87cm]{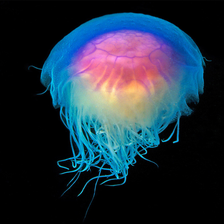}}
    \subfigure[Blue]{\includegraphics[height=2cm, width=2.87cm]{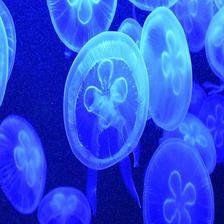}}
    \subfigure[Blue]{\includegraphics[height=2cm, width=2.87cm]{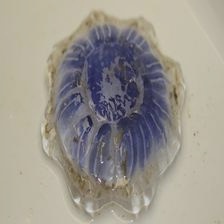}}
    \subfigure[Compass]{\includegraphics[height=2cm, width=2.87cm]{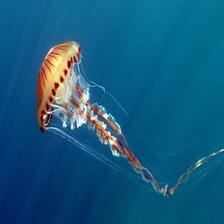}}
    \subfigure[Compass]{\includegraphics[height=2cm, width=2.87cm]{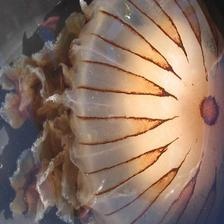}}
    \subfigure[Compass]{\includegraphics[height=2cm, width=2.87cm]{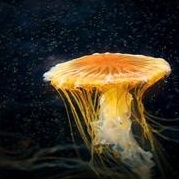}}
    \subfigure[Lions Mane]{\includegraphics[height=2cm, width=2.87cm]{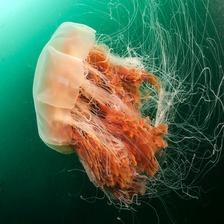}}
    \subfigure[Lions Mane]{\includegraphics[height=2cm, width=2.87cm]{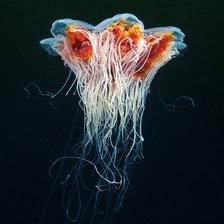}}
    \subfigure[Lions Mane]{\includegraphics[height=2cm, width=2.87cm]{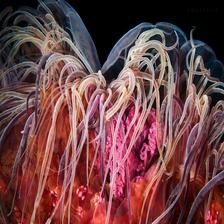}}
    \subfigure[Mauve Stinger]{\includegraphics[height=2cm, width=2.87cm]{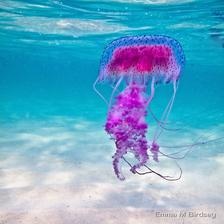}}
    \subfigure[Mauve Stinger]{\includegraphics[height=2cm, width=2.87cm]{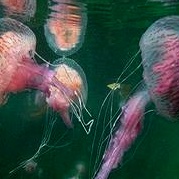}}
    \subfigure[Mauve Stinger]{\includegraphics[height=2cm, width=2.87cm]{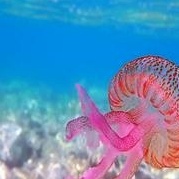}}
    \subfigure[Moon]{\includegraphics[height=2cm, width=2.87cm]{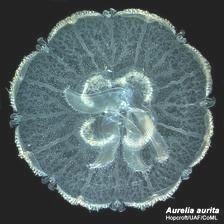}}
    \subfigure[Moon]{\includegraphics[height=2cm, width=2.87cm]{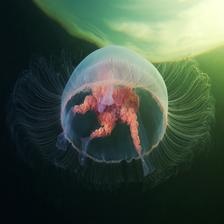}}
    \subfigure[Moon]{\includegraphics[height=2cm, width=2.87cm]{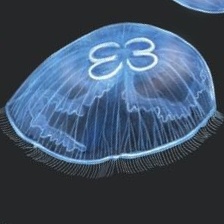}}
    \vspace{-2mm}
    \caption{Sample images of different jellyfish species from the jellyfish image dataset, including Barrel jellyfish, Blue jellyfish, Compass jellyfish, Lion’s Mane jellyfish, Mauve Stinger jellyfish, and Moon jellyfish, illustrating the diversity in color, shape, and texture. \vspace{-3mm}}
    \label{fig:jfsh}
\end{figure*}

\subsection{Dataset Description}
The dataset used in this research, “Jellyfish Image Dataset," was collected from Kaggle \cite{dataset}. It comprises high-quality underwater images of jellyfish, captured in various underwater conditions. The dataset includes six distinct jellyfish species: Barrel Jellyfish, Blue Jellyfish, Compass Jellyfish, Lion’s Mane Jellyfish, Mauve Stinger Jellyfish, and Moon Jellyfish. For each species, the dataset consists 150 images for training, 15 for testing, and 15 for validation, resulting in a total of 1,080 images across six classes. Each image is meticulously labeled with the corresponding species, ensuring accurate and effective supervised learning. Three Samples from each jellyfish class are displayed in Figure \ref{fig:jfsh}

\subsection{Data Preprocessing}
The preprocessing pipeline for the dataset involved multiple techniques to ensure optimal input quality for the model. First, normalization was applied by rescaling pixel values from the original range [0, 255] to [0, 1], which standardizes the data and accelerates the training process. Additionally, images were resized to a uniform dimension (224x224 pixels) to maintain consistency across the dataset and meet the input requirements of the model. The preprocessing was implemented using dataset mapping with TensorFlow's map() function, which applies transformations efficiently to each data sample. To enhance computational efficiency, parallelized data preprocessing was enabled using the AUTOTUNE parameter, ensuring that preprocessing and data loading were optimized for the available hardware resources.

\subsection{Data Augmentation}
To enhance the diversity and robustness of the dataset, data augmentation techniques were employed to artificially expand the training data by generating 10,000 additional samples. The augmentation process included random transformations such as RandomFlip, which flips images horizontally to simulate different viewing angles, and RandomRotation, which rotates images randomly within a specified range to account for variations in orientation. Additionally, RandomZoom was applied to simulate varying distances of objects, while RandomWidth and RandomHeight adjustments were used to alter the aspect ratio, mimicking real-world scenarios of object distortion. These transformations not only increased the dataset size but also helped the model generalize better by exposing it to a wide range of image variations.

\subsection{Deep Learning Architectures}
\vspace{-1mm}In this research, we employed several convolutional neural network (CNN) architectures to extract features and classify different species of jellyfish from underewater images. Specifically, we utilized MobileNetV3 \cite{mobilenetv3}, ResNet50 \cite{resnet}, EfficientNetV2-B0 \cite{efficientnet}, and VGG16 \cite{vgg16} to leverage their state-of-the-art performance in image analysis. Each model was initially trained as a standalone classifier to directly identify jellyfish species. Subsequently, we used these models as feature extractors, where the extracted features were fed into machine learning and multilayer perceptron classifiers for further classification.  \vspace{2mm}

\subsubsection{CNN-Based Classification}
In this approach, we utilized CNNs to perform end-to-end learning, where the network simultaneously learns to extract features and classify jellyfish species. The raw input images were processed through a series of convolutional layers that identified key visual patterns, such as edges, textures, and shapes, which are critical for classification. These layers were followed by pooling operations to reduce dimensionality while retaining essential features. The feature maps generated were then flattened and passed through fully connected dense layers. To output class probabilities, the final dense layer employed a softmax activation function, ensuring that each jellyfish species was assigned a probability score. The species with the highest probability was selected as the final prediction. \vspace{2mm}

\subsubsection{CNN-Based Feature Extraction}
For feature extraction, we leveraged pre-trained CNN architectures to utilize their learned hierarchical representations of visual features. These models learns hierarchical features from input images, capturing essential details such as edges, textures, and more complex patterns through successive convolution and pooling layers. Instead of using the fully connected layers for classification, we extracted the intermediate feature representations from the convolutional layers. These high-dimensional features were flattened and used as input to external ML or FNN classifiers, such as Support Vector Machine (SVM) or Artificial Neural Network (ANN), which performed the final classification task.

\subsection{Model Selection}
To identify different species of jellyfish, we evaluated the performance of both traditional machine learning classifiers \cite{maternal, strawberry} and advanced feedforward neural network (FNN) classifiers \cite{mlp} \cite{hossen2025july}. The features extracted from pre-trained CNN architectures were used as input for these classifiers, enabling robust and efficient species identification. \vspace{3mm}

\subsubsection{Machine Learning Classifiers}
We employed several machine learning classifiers to analyze the extracted features and classify jellyfish species. These included Support Vector Machine (SVM), Random Forest (RF), Decision Tree (DT), Logistic Regression (LR), Gradient Boosting (GB), Extreme Gradient Boosting (XGBoost), and Light Gradient Boosting Machine (LightGBM). Each of these classifiers offers distinct advantages in handling structured data and high-dimensional feature spaces. SVM excels in creating decision boundaries for complex datasets, while RF and DT are well-suited for capturing non-linear patterns. LR provides a simple yet effective baseline for binary and multiclass classification, and ensemble-based methods like GB, XGBoost, and LightGBM are known for their robustness and accuracy. \vspace{3mm}

\subsubsection{Feedforward Neural Network Classifiers}
In addition to traditional ML classifiers, we utilized advanced FNN classifiers to further enhance classification performance. These included an Artificial Neural Network (ANN), a Radial Basis Function Neural Network (RBFNN) with an SVM RBF kernel, and an Autoencoder-based model. The ANN consisted of multiple hidden layers with activation functions to model non-linear relationships in the feature space effectively. The RBFNN leveraged the SVM RBF kernel to capture complex decision boundaries in the feature space, combining the strengths of neural networks and kernel-based methods. The Autoencoder-based model employed an unsupervised approach to learn compressed representations of features, which were then fine-tuned for classification tasks. These FNN-based approaches allowed for greater flexibility in learning intricate patterns in the extracted features and contributed to achieving state-of-the-art accuracy. \vspace{-3mm}

\section{EXPERIMENTAL SETUP \& IMPLEMENTATION}
\vspace{-2mm}
\subsection{Experimental Setup}
The experiments were conducted on a Windows 11 system (64-bit) with an Intel Core i5 8th Gen CPU (1.60–3.90 GHz), 12 GB DDR4 RAM, and an NVIDIA MX250 GPU. Jupyter Notebook, managed via Anaconda Navigator, was used for implementation, including image processing and model training. The training and evaluation process took over ten hours, highlighting its computational demands.

\subsection{Implementation Details}
The implementation process for identifying jellyfish species follows a structured and modular approach. The dataset is first sourced from a public repository and preprocessed through several techniques to ensure uniformity. Data augmentation techniques, such as random flipping, rotation, cropping, etc., are then applied to enhance the dataset and increase model robustness. The classification process is divided into three distinct methodologies. \vspace{2mm}

\noindent Firstly, four CNN architectures—MobileNetV3, ResNet50, EfficientNetV2-B0, and VGG16—are employed for directly classifying jellyfish species, utilizing the softmax function for prediction. Secondly, features are extracted from images using the same CNN models, and these features are subsequently fed into seven machine learning classifiers to perform classification. Lastly, three Feedforward Neural Network (FNN) models were utilized: an Artificial Neural Network (ANN), a Radial Basis Function Neural Network (RBFNN) based on the SVM kernel, and an autoencoder-based model. 

\noindent Each model's performance is evaluated using metrics such as accuracy, precision, recall, and F1-score, enabling a comprehensive comparison to identify the most effective model for jellyfish species classification. The following Algorithm \ref{alg:jellyfish} illustrates the overall jellyfish classification procedure.

\begin{algorithm}[]
\scriptsize
\caption{Algorithm for Jellyfish Species Classification}
\label{alg:jellyfish}

\textbf{Initialize:} $DS$, $PP$, $DA$, $CNN$, $FE$, $ML$, $FNN$ \tcp*[r]{Initialize components}

$DS \gets \text{Jellyfish Image Dataset (6 Jellyfish Species)}$ \; 

$PP \gets [\text{Resizing, Rescaling, Data Mapping, AUTOTUNE}]$ \; 

$DA \gets [\text{Random Flip, rotation, zoom, height, width}]$ \tcp*[r]{Data Augmentation (Generate 10,000 additional samples)}
$CNN \gets [\text{MobileNetV3, VGG16, ResNet50, EfficientNetV2-B0}]$ \tcp*[r]{Pre-trained CNN Models}
$FE \gets \text{Feature Extraction (Removing Classifier Head)}$ \; 

$ML \gets [\text{SVM, RF, DT, LR, GB, XGB, LGBM}]$ \; 

$FNN \gets [\text{ANN, RBFNN, Autoencoder base}]$ \;

$X \gets DS[\text{Images}], \; Y \gets DS[\text{Labels}]$ \;

\For{$i \gets 1$ \KwTo $|CNN|$}{
    \tcp{Direct Classification with CNN}
    $CNN[i].\text{add}(\text{Softmax Function})$ \;
    $CNN[i].\text{compile}(\text{optimizer=Adam})$ \;
    $CNN[i].\text{fit}(X, Y, \text{epochs=20, batch\_size=32})$ \;
    
    $\text{Evaluate } \to [\text{Accuracy, Precision, Recall, F1-Score}]$ \;

    \tcp{Feature Extraction and ML Classification}
    $FV \gets FE(CNN[i], X)$ \tcp*[r]{Extract Features}
    \For{$j \gets 1$ \KwTo $|ML|$}{
        $X_T, Y_T, x_t, y_t \gets \text{Train-Test Split (70:30)}(FV, Y)$ \;
        $ML[j].\text{fit}(X_T, Y_T)$ \tcp*[r]{Train Classifier}
        $y_{\text{pred}} \gets ML[j].\text{predict}(x_t)$ \tcp*[r]{Test Classifier}
        $\text{Evaluate } \to [\text{Accuracy, Precision, Recall, F1-Score}]$ \;
    }

    \tcp{Feature Extraction and FNN Classification}
    \For{$k \gets 1$ \KwTo $|FNN|$}{
        $X_T, Y_T, x_t, y_t \gets \text{Train-Test Split (70:30)}(FV, Y)$ \;
        $FNN[k].\text{fit}(X_T, Y_T)$ \tcp*[r]{Train FNN}
        $y_{\text{pred}} \gets FNN[k].\text{predict}(x_t)$ \tcp*[r]{Test FNN}
        $\text{Evaluate } \to [\text{Accuracy, Precision, Recall, F1-Score}]$ \;
    }
}
\textbf{Deinitialize:} $DS$, $PP$, $DA$, $CNN$, $FE$, $ML$, $FNN$ \tcp*[r]{Shutdown} 
\end{algorithm} \vspace{-3mm}
\vspace{-3mm}
\section{RESULT ANALYSIS \& DISCUSSION}
\vspace{-2mm} This section provides a detailed evaluation of various models and methods used for jellyfish species classification. The performance of the models is assessed using metrics such as Accuracy (A), Precision (P), Recall (R), and F1-score (F1) to determine their effectiveness in distinguishing between jellyfish species. The analysis includes comparisons of CNN architectures, hybrid machine learning, and multilayer perceptron classifiers, highlighting their strengths and limitations. The results are presented through detailed tables and visualizations, offering a clear perspective on the most efficient techniques for jellyfish detection.

\subsection{Result Analysis}
The table \ref{tab:sm} presents the performance of four convolutional neural network (CNN) architectures—VGG16, MobileNetV3, ResNet50, and EfficientNetV2-B0—evaluated for jellyfish species classification using Softmax activation. The MobileNetV3 achieved the highest performance across all metrics, with 93\% accuracy, precision, recall, and F1-score. ResNet50 and EfficientNet-B0 both exhibited similar performance, scoring 92\% across all metrics. VGG16 showed the lowest performance among the models, achieved 89\% for each metric. ResNet50 and EfficientNet-B0 both exhibited similar performance, scoring 92\% across all metrics. VGG16 showed the lowest performance among the models, achieved 89\% for each metric. \vspace{-3mm}
\begin{table}[htbp]
\scriptsize
\setlength{\tabcolsep}{3.8pt}
\renewcommand{\arraystretch}{1}
\centering
\caption{Performance of CNN Architectures with Softmax Activation \vspace{-2mm}}
\label{tab:sm}
\begin{tabular}{|c|c|c|c|c|}
\hline
\textbf{Architecture/Model} & \textbf{Accuracy(\%)} & \textbf{Precision(\%)} & \textbf{Recall(\%)} & \textbf{F1-score(\%)} \\ \hline
VGG16 & 89.0 & 89.0 & 89.0 & 89.0 \\ \hline
MobileNetV3 & 93.0 & 93.0 & 93.0 & 93.0 \\ \hline
ResNet50 & 92.0 & 91.0 & 91.0 & 91.0 \\ \hline
EfficientNetV2-B0 & 92.0 & 92.0 & 92.0 & 92.0 \\ \hline
\end{tabular}
\end{table}
\begin{table*}[]
\scriptsize
\setlength{\tabcolsep}{5.7pt}
\renewcommand{\arraystretch}{1}
\caption{Performance of Machine Learning Classifiers Using Features Extracted by CNN Architectures \vspace{-2mm}}
\centering
\label{tab:ml}
\begin{tabular}{|c|cccccccccccccccc|}
\hline
\multirow{3}{*}{\textbf{\begin{tabular}[c]{@{}c@{}}ML\\ Classifier\end{tabular}}} & \multicolumn{16}{c|}{\textbf{Feature Extraction Techniques}} \\ \cline{2-17} 
 & \multicolumn{4}{c|}{\textbf{VGG16}} & \multicolumn{4}{c|}{\textbf{MobileNetV3}} & \multicolumn{4}{c|}{\textbf{ResNet50}} & \multicolumn{4}{c|}{\textbf{EfficientNetV2-B0}} \\ \cline{2-17} 
 & \multicolumn{1}{c|}{\textbf{A(\%)}} & \multicolumn{1}{c|}{\textbf{P(\%)}} & \multicolumn{1}{c|}{\textbf{R(\%)}} & \multicolumn{1}{c|}{\textbf{F1(\%)}} & \multicolumn{1}{c|}{\textbf{A(\%)}} & \multicolumn{1}{c|}{\textbf{P(\%)}} & \multicolumn{1}{c|}{\textbf{R(\%)}} & \multicolumn{1}{c|}{\textbf{F1(\%)}} & \multicolumn{1}{c|}{\textbf{A(\%)}} & \multicolumn{1}{c|}{\textbf{P(\%)}} & \multicolumn{1}{c|}{\textbf{R(\%)}} & \multicolumn{1}{c|}{\textbf{F1(\%)}} & \multicolumn{1}{c|}{\textbf{A(\%)}} & \multicolumn{1}{c|}{\textbf{P(\%)}} & \multicolumn{1}{c|}{\textbf{R(\%)}} & \textbf{F1(\%)} \\ \hline
SVM & \multicolumn{1}{c|}{87.0} & \multicolumn{1}{c|}{86.0} & \multicolumn{1}{c|}{89.0} & \multicolumn{1}{c|}{87.0} & \multicolumn{1}{c|}{97.0} & \multicolumn{1}{c|}{97.0} & \multicolumn{1}{c|}{97.0} & \multicolumn{1}{c|}{97.0} & \multicolumn{1}{c|}{96.0} & \multicolumn{1}{c|}{96.0} & \multicolumn{1}{c|}{96.0} & \multicolumn{1}{c|}{96.0} & \multicolumn{1}{c|}{90.0} & \multicolumn{1}{c|}{90.0} & \multicolumn{1}{c|}{92.0} & 90.0 \\ \hline
RF & \multicolumn{1}{c|}{81.0} & \multicolumn{1}{c|}{81.0} & \multicolumn{1}{c|}{84.0} & \multicolumn{1}{c|}{81.0} & \multicolumn{1}{c|}{83.0} & \multicolumn{1}{c|}{83.0} & \multicolumn{1}{c|}{86.0} & \multicolumn{1}{c|}{84.0} & \multicolumn{1}{c|}{88.0} & \multicolumn{1}{c|}{88.0} & \multicolumn{1}{c|}{88.0} & \multicolumn{1}{c|}{88.0} & \multicolumn{1}{c|}{79.0} & \multicolumn{1}{c|}{81.0} & \multicolumn{1}{c|}{83.0} & 80.0 \\ \hline
DT & \multicolumn{1}{c|}{59.0} & \multicolumn{1}{c|}{59.0} & \multicolumn{1}{c|}{61.0} & \multicolumn{1}{c|}{59.0} & \multicolumn{1}{c|}{60.0} & \multicolumn{1}{c|}{61.0} & \multicolumn{1}{c|}{63.0} & \multicolumn{1}{c|}{61.0} & \multicolumn{1}{c|}{60.0} & \multicolumn{1}{c|}{60.0} & \multicolumn{1}{c|}{61.0} & \multicolumn{1}{c|}{60.0} & \multicolumn{1}{c|}{57.0} & \multicolumn{1}{c|}{57.0} & \multicolumn{1}{c|}{59.0} & 57.0 \\ \hline
LR & \multicolumn{1}{c|}{88.0} & \multicolumn{1}{c|}{88.0} & \multicolumn{1}{c|}{90.0} & \multicolumn{1}{c|}{88.0} & \multicolumn{1}{c|}{96.0} & \multicolumn{1}{c|}{97.0} & \multicolumn{1}{c|}{96.0} & \multicolumn{1}{c|}{96.0} & \multicolumn{1}{c|}{92.0} & \multicolumn{1}{c|}{91.0} & \multicolumn{1}{c|}{91.0} & \multicolumn{1}{c|}{91.0} & \multicolumn{1}{c|}{88.0} & \multicolumn{1}{c|}{88.0} & \multicolumn{1}{c|}{90.0} & 89.0 \\ \hline
GB & \multicolumn{1}{c|}{80.0} & \multicolumn{1}{c|}{81.0} & \multicolumn{1}{c|}{83.0} & \multicolumn{1}{c|}{81.0} & \multicolumn{1}{c|}{88.0} & \multicolumn{1}{c|}{88.0} & \multicolumn{1}{c|}{89.0} & \multicolumn{1}{c|}{89.0} & \multicolumn{1}{c|}{86.0} & \multicolumn{1}{c|}{85.0} & \multicolumn{1}{c|}{86.0} & \multicolumn{1}{c|}{85.0} & \multicolumn{1}{c|}{77.0} & \multicolumn{1}{c|}{79.0} & \multicolumn{1}{c|}{79.0} & 78.0 \\ \hline
XGB & \multicolumn{1}{c|}{76.0} & \multicolumn{1}{c|}{76.0} & \multicolumn{1}{c|}{79.0} & \multicolumn{1}{c|}{76.0} & \multicolumn{1}{c|}{88.0} & \multicolumn{1}{c|}{87.0} & \multicolumn{1}{c|}{89.0} & \multicolumn{1}{c|}{88.0} & \multicolumn{1}{c|}{89.0} & \multicolumn{1}{c|}{88.0} & \multicolumn{1}{c|}{88.0} & \multicolumn{1}{c|}{88.0} & \multicolumn{1}{c|}{77.0} & \multicolumn{1}{c|}{79.0} & \multicolumn{1}{c|}{82.0} & 78.0 \\ \hline
LGBM & \multicolumn{1}{c|}{80.0} & \multicolumn{1}{c|}{80.0} & \multicolumn{1}{c|}{84.0} & \multicolumn{1}{c|}{80.0} & \multicolumn{1}{c|}{88.0} & \multicolumn{1}{c|}{88.0} & \multicolumn{1}{c|}{91.0} & \multicolumn{1}{c|}{88.0} & \multicolumn{1}{c|}{87.0} & \multicolumn{1}{c|}{88.0} & \multicolumn{1}{c|}{91.0} & \multicolumn{1}{c|}{88.0} & \multicolumn{1}{c|}{76.0} & \multicolumn{1}{c|}{79.0} & \multicolumn{1}{c|}{81.0} & 78.0 \\ \hline
\end{tabular} \vspace{-2mm}
\end{table*}
\begin{table*}[]
\scriptsize
\setlength{\tabcolsep}{5.5pt}
\renewcommand{\arraystretch}{1}
\caption{Performance of Feedforward Neural Network Classifiers Using Features Extracted by CNN Architectures \vspace{-2mm}}
\centering
\label{tab:mlp}
\begin{tabular}{|c|cccccccccccccccc|}
\hline
\multirow{3}{*}{\textbf{\begin{tabular}[c]{@{}c@{}}MLP\\ Classifier\end{tabular}}} & \multicolumn{16}{c|}{\textbf{Feature Extraction Techniques}} \\ \cline{2-17} 
 & \multicolumn{4}{c|}{\textbf{VGG16}} & \multicolumn{4}{c|}{\textbf{MobileNetV3}} & \multicolumn{4}{c|}{\textbf{ResNet50}} & \multicolumn{4}{c|}{\textbf{EfficientNet-B0}} \\ \cline{2-17} 
 & \multicolumn{1}{c|}{\textbf{A(\%)}} & \multicolumn{1}{c|}{\textbf{P(\%)}} & \multicolumn{1}{c|}{\textbf{R(\%)}} & \multicolumn{1}{c|}{\textbf{F1(\%)}} & \multicolumn{1}{c|}{\textbf{A(\%)}} & \multicolumn{1}{c|}{\textbf{P(\%)}} & \multicolumn{1}{c|}{\textbf{R(\%)}} & \multicolumn{1}{c|}{\textbf{F1(\%)}} & \multicolumn{1}{c|}{\textbf{A(\%)}} & \multicolumn{1}{c|}{\textbf{P(\%)}} & \multicolumn{1}{c|}{\textbf{R(\%)}} & \multicolumn{1}{c|}{\textbf{F1(\%)}} & \multicolumn{1}{c|}{\textbf{A(\%)}} & \multicolumn{1}{c|}{\textbf{P(\%)}} & \multicolumn{1}{c|}{\textbf{R(\%)}} & \textbf{F1(\%)} \\ \hline
ANN & \multicolumn{1}{c|}{86.0} & \multicolumn{1}{c|}{86.0} & \multicolumn{1}{c|}{88.0} & \multicolumn{1}{c|}{87.0} & \multicolumn{1}{c|}{98.0} & \multicolumn{1}{c|}{98.0} & \multicolumn{1}{c|}{98.0} & \multicolumn{1}{c|}{98.0} & \multicolumn{1}{c|}{90.0} & \multicolumn{1}{c|}{90.0} & \multicolumn{1}{c|}{92.0} & \multicolumn{1}{c|}{91.0} & \multicolumn{1}{c|}{89.0} & \multicolumn{1}{c|}{89.0} & \multicolumn{1}{c|}{91.0} & 89.0 \\ \hline
\begin{tabular}[c]{@{}c@{}}RBFNN\end{tabular} & \multicolumn{1}{c|}{85.0} & \multicolumn{1}{c|}{85.0} & \multicolumn{1}{c|}{88.0} & \multicolumn{1}{c|}{86.0} & \multicolumn{1}{c|}{90.0} & \multicolumn{1}{c|}{90.0} & \multicolumn{1}{c|}{91.0} & \multicolumn{1}{c|}{90.0} & \multicolumn{1}{c|}{89.0} & \multicolumn{1}{c|}{89.0} & \multicolumn{1}{c|}{93.0} & \multicolumn{1}{c|}{90.0} & \multicolumn{1}{c|}{87.0} & \multicolumn{1}{c|}{87.0} & \multicolumn{1}{c|}{89.0} & 88.0 \\ \hline
Autoencoder & \multicolumn{1}{c|}{81.0} & \multicolumn{1}{c|}{81.0} & \multicolumn{1}{c|}{84.0} & \multicolumn{1}{c|}{82.0} & \multicolumn{1}{c|}{96.0} & \multicolumn{1}{c|}{97.0} & \multicolumn{1}{c|}{96.0} & \multicolumn{1}{c|}{97.0} & \multicolumn{1}{c|}{91.0} & \multicolumn{1}{c|}{91.0} & \multicolumn{1}{c|}{94.0} & \multicolumn{1}{c|}{91.0} & \multicolumn{1}{c|}{89.0} & \multicolumn{1}{c|}{89.0} & \multicolumn{1}{c|}{90.0} & 89.0 \\ \hline
\end{tabular} \vspace{-2mm}
\end{table*}
\begin{figure*}[htbp]
    \centering
    \subfigure[]{\label{fig:a}\includegraphics[height=4.8cm, width=5.85cm]{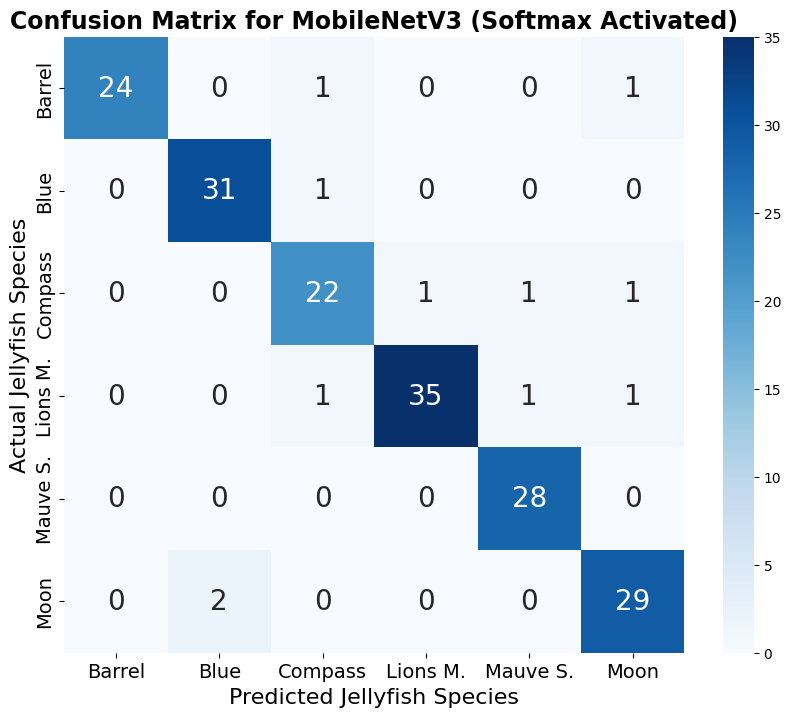}}
    \subfigure[]{\label{fig:b}\includegraphics[height=4.8cm, width=5.85cm]{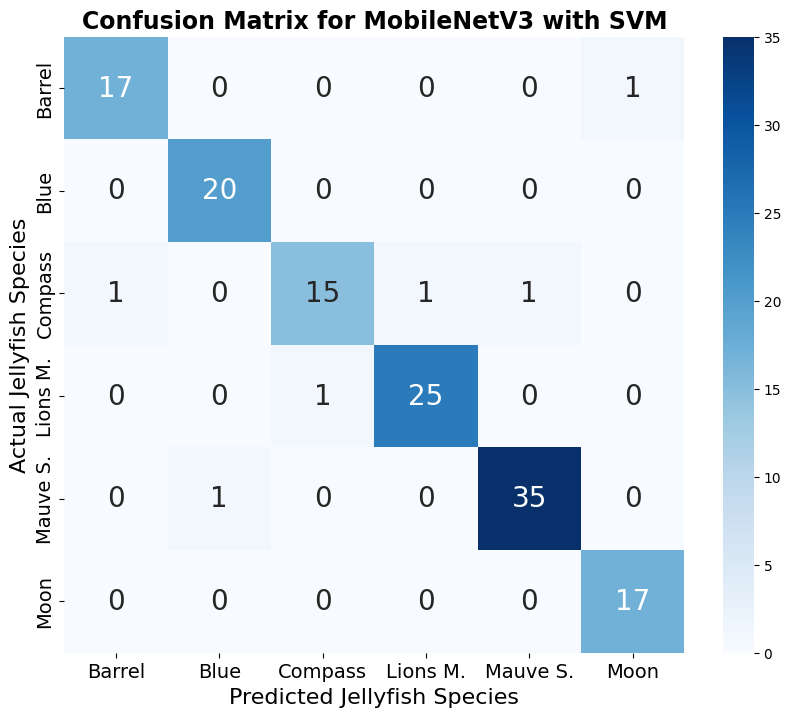}}
    \subfigure[]{\label{fig:c}\includegraphics[height=4.8cm, width=5.85cm]{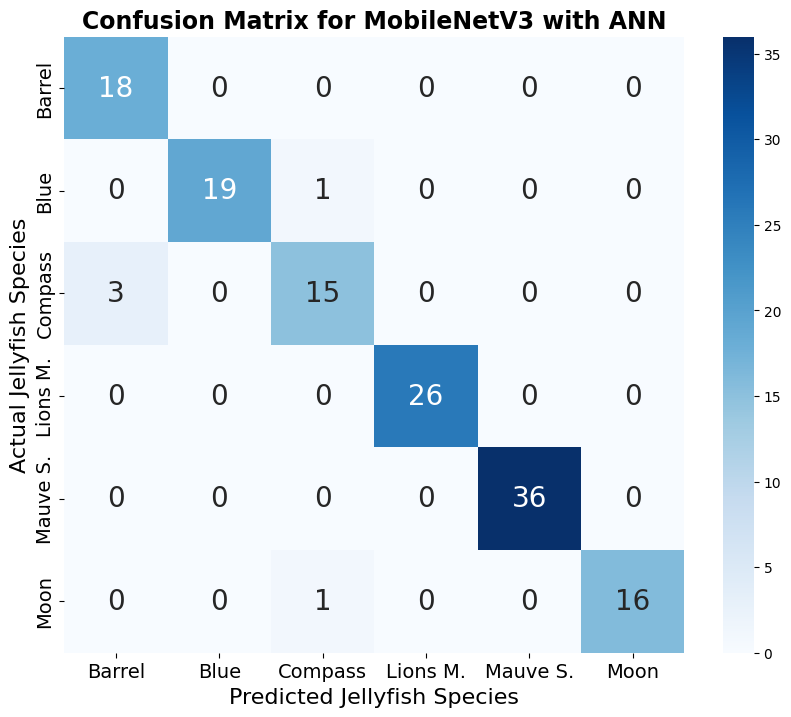}}
    \vspace{-0.5cm}
    \caption{Confusion matrices comparing the performance of jellyfish species classification using MobileNetV3 with three different approaches: (a) softmax activation, (b) Feature Extraction combined with Support Vector Machine, and (c) Feature Extraction Combined with Artificial Neural Network. \vspace{-2mm}}
    \label{fig:cm}
\end{figure*}
\begin{figure*}[htbp]
    \centering
    \subfigure[]{\label{fig:e}\includegraphics[height=3.5cm, width=5.85cm]{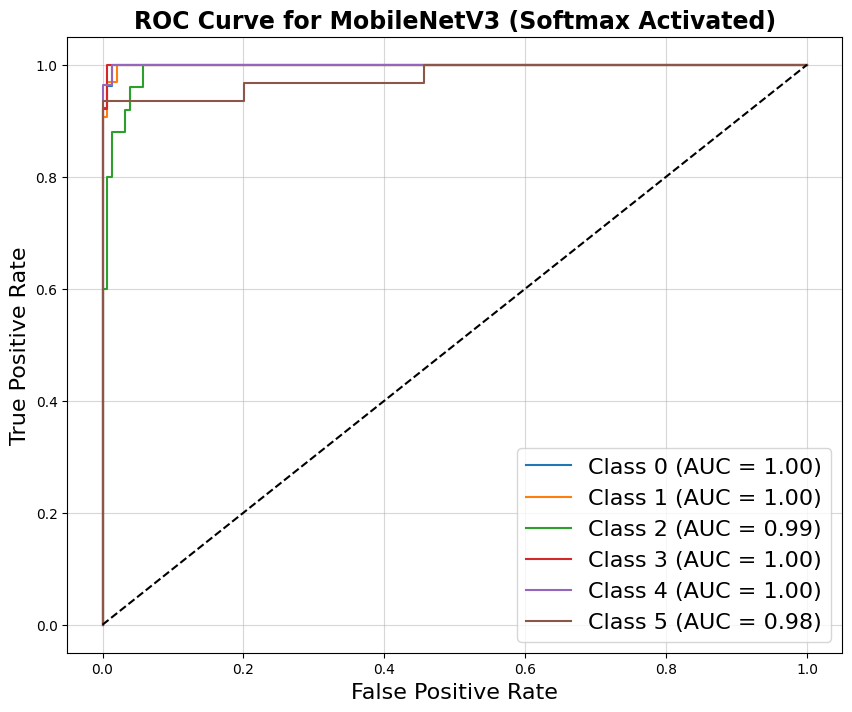}}
    \subfigure[]{\label{fig:f}\includegraphics[height=3.5cm, width=5.85cm]{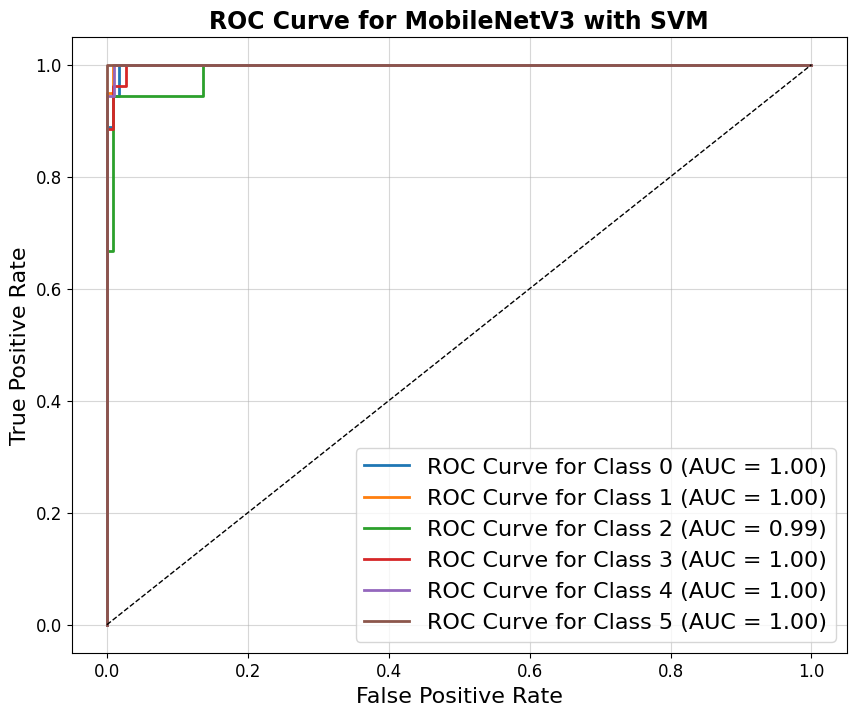}}
    \subfigure[]{\label{fig:g}\includegraphics[height=3.5cm, width=5.85cm]{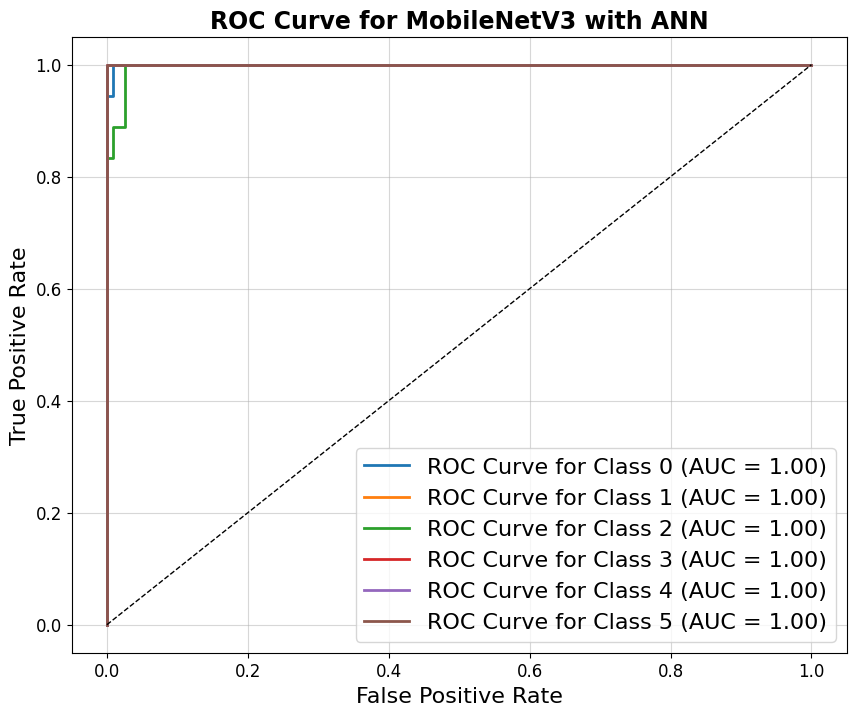}}
    \vspace{-0.5cm}
    \caption{Receiver-operating characteristic (ROC) curve comparing the performance of jellyfish species classification using MobileNetV3 with three different approaches: (a) softmax activation, (b) Feature Extraction combined with Support Vector Machine, and (c) Feature Extraction Combined with Artificial Neural Network. The figures show the ROC Curve across six classes: 0, 1, 2, 3, 4, and 5 with their values.} \vspace{-3mm}
    \label{fig:rc}
\end{figure*}

The table \ref{tab:ml} presents the performance of seven machine learning classifiers using features extracted by four convolutional neural network architectures. MobileNetV3 combined with SVM achieved the best overall results of 97\% across all metrics. ResNet50 also showed strong results, with SVM achieved 96\% Accuracy. In contrast, Decision Tree demonstrated the weakest performance across all CNN architectures, with Accuracy values as low as 59\% when paired with VGG16 and ResNet50. Overall, MobileNetV3 outperformed other feature extraction techniques across most classifiers, while EfficientNet-B0 provided competitive but slightly lower results, particularly with RF and LGBM classifiers.  %\vspace{2mm}

\noindent The table \ref{tab:mlp} illustrates the performance of three Feedforward Neural Network (FNN) classifiers on features extracted from four CNN architectures. The Artificial Neural Network paired with MobileNetV3 achieved the best performance across all metrics, recording an Accuracy, Precision, Recall, and F1-score of 98\%. RBFNN and Autoencoder based classifier also demonstrated strong results when combined with MobileNetV3, with both achieving an Accuracy of 90\% and 96\%, respectively. ResNet50 with ANN and Autoencoder base model also demonstrates high Accuracy 90\% and 91\%, respectively, reflecting its strong predictive capabilities. In contrast, VGG16 produced comparatively lower results, with the Autoencoder achieving an Accuracy of 81\%. Overall, MobileNetV3 proved to be the most effective feature extractor for all Feedforward Neural Network classifiers. %while ResNet50 also delivered competitive performance, particularly with RBFNN. 

\noindent Fig. \ref{fig:cm} illustrates the confusion matrices of different classification approaches for jellyfish species detection using MobileNetV3. The softmax-activated MobileNetV3 model achieves strong classification accuracy, correctly identifying most species with minimal misclassification. The feature extraction-based Support Vector Machine and Artificial Neural Network models further enhance performance, reducing misclassification rates. Notably, the Artificial Neural Network approach achieves the highest consistency, with near-perfect classification in most cases. \vspace{2mm}

\noindent Fig. \ref{fig:rc} displays the ROC curves demonstrate the performance of MobileNetV3 in jellyfish species classification. All models exhibit excellent classification capabilities, achieving near-perfect Area Under the Curve (AUC) values across all six classes. The Artificial Neural Network based approach achieved the highest overall effectiveness. The ANN model consistently attains an AUC of 1.00 across all classes, demonstrating its superior discriminatory power. While the SVM and softmax-activated models also perform well, with most AUC values close to 1.00, Artificial Neural Networks stands out as the best model for accurate jellyfish species classification.

\subsection{Discussion \& Limitations}
In this study, we explored various machine learning models, evaluating them through multiple performance metrics, and the results show Artificial Neural Network combined with MobileNetV3 consistently outperformed other approaches. With its superior accuracy, precision, recall, F1 score, and AUC scores, this model demonstrated strong classification capabilities, making it the optimal choice for automated jellyfish detection. However, this study has certain limitations. The dataset used is relatively small, which may affect the model’s ability to generalize to a wider range of jellyfish species and varying environmental conditions. Additionally, our approach relies solely on image-based features, whereas integrating ecological factors such as water temperature, salinity, and seasonal variations could further improve classification accuracy. Another challenge is the computational cost associated with deep learning models, which may hinder real-time deployment in low-resource settings. \vspace{-3mm}

\section{CONCLUSION \& FUTURE WORK}
\vspace{-2mm} Accurate identification of jellyfish species is crucial for marine biodiversity monitoring and ecosystem management. This study demonstrates the potential of AI-driven classification in marine biodiversity research by effectively distinguishing jellyfish species through deep learning and hybrid machine learning approaches. By leveraging CNN-based feature extraction and integrating it with advanced classifiers, the proposed models achieve high classification performance, with the Artificial Neural Network combined with MobileNetV3 attaining the highest accuracy of 98\%, along with superior precision, recall, and F1-score. These findings contribute to marine species identification research, highlighting the potential of AI-driven classification in biodiversity studies and ecological monitoring. Future work can focus on expanding the dataset to include a wider range of jellyfish species, enhancing real-time classification efficiency, and incorporating additional ecological factors such as habitat conditions and seasonal variations to improve model robustness. Moreover, exploring advanced deep learning techniques, including transformer-based vision models and self-supervised learning, can further refine classification accuracy and enhance the model’s generalizability. \vspace{-3mm}

\bibliographystyle{ieeetr}
\bibliography{ref.bib}
\end{document}